%% file: generative-concatenative.tex
\documentclass{article}

\usepackage{times}
\usepackage{graphicx} 

\usepackage[round]{natbib}

\usepackage{booktabs}

\usepackage{algorithm}
\usepackage{algorithmic}
\usepackage{amsthm}
\usepackage{amsmath}
\usepackage{amsfonts}
\usepackage{bbm}
\usepackage{titlesec}
\usepackage{caption}
\usepackage{subcaption}
\usepackage[colorlinks=true,allcolors=black]{hyperref}
\usepackage{csquotes}
\usepackage{appendix}
\usepackage{subcaption}
\usepackage{url}
\usepackage{tabularx}
\usepackage{hhline}
\usepackage{multirow}
\usepackage[usenames, dvipsnames]{color}
\definecolor{myorange}{RGB}{255, 131, 0}

\DeclareMathOperator*{\argmax}{\mathrm{argmax}}

\usepackage{listings}

\usepackage{uai} 

\title{Generative Concatenative Nets Jointly Learn to Write and Classify Reviews}

\author{%
Zachary C. Lipton\thanks{Author website: http://zacklipton.com }, \enskip Sharad Vikram \thanks{Author website: http://www.sharadvikram.com}, \enskip Julian McAuley \thanks{Author website: http://cseweb.ucsd.edu/$\sim$jmcauley/}\\
Computer Science and Engineering\\
University of California, San Diego\\
La Jolla, CA 92093, USA \\
\texttt{\{zlipton,svikram,jmcauley\}@cs.ucsd.edu} \\
}

\begin{document} 
\maketitle

\begin{abstract} 
\input{sections/abstract.tex}

\end{abstract} 

\section{Introduction}
\input{sections/introduction.tex}

\subsection{Contributions}
\input{sections/contributions.tex}

\section{Data}
\input{sections/data.tex}

\section{Recurrent Neural Network Methodology}
\input{sections/recurrent-nets.tex}


\section{Experiments}
\input{sections/experiments.tex}

\section{Related Work}
\input{sections/related.tex}

\section{Conclusion}
\input{sections/discussion.tex}

\bibliography{gcn.bib}
\bibliographystyle{plain}









\end{document}

%% file: sections/abstract.tex
A recommender system's basic task 
is to estimate how users will respond to unseen items. This is typically modeled in terms 
of how a user might \emph{rate} a product, 
but here we aim to extend such approaches to model 
how a user would \emph{write} about the product.
To do so, we design a character-level Recurrent Neural Network (RNN)
that generates personalized product reviews.
The network convincingly learns styles and opinions 
of nearly 1000 distinct authors, 
using a large corpus of reviews from \emph{BeerAdvocate.com}.
It also tailors reviews to describe specific items, categories, and star ratings.
Using a simple input replication strategy,
the Generative Concatenative Network (GCN) 
preserves the signal of static auxiliary inputs 
across wide sequence intervals.
Without any additional training,
the generative model can classify reviews,
identifying the author of the review,
the product category, 
and the sentiment (rating), 
with remarkable accuracy.
Our evaluation shows the GCN 
captures complex dynamics in text, 
such as the effect of negation, misspellings, slang, 
and large vocabularies gracefully
absent any machinery explicitly dedicated to the purpose.

%% file: sections/introduction.tex
Recommender systems assist users 
in navigating an unprecedented selection of items, 
personalizing services to a diverse set of users with distinct individual tastes. 
Typical approaches surface items 
that a customer is likely to purchase or rate highly,
providing a basic set of primitives 
for building functioning internet applications. 
Our goal, however, is to create richer user experiences,
not only recommending products 
but generating personalized descriptive text.
Engaged users may wish to know what precisely 
their impression of an item is expected to be, 
not simply whether it warrants a thumbs up or thumbs down.
\emph{Customer reviews} help this issue to some extent,
but large volumes of reviews are difficult to sift through, 
especially for users interested in some niche aspect.
In this work, we address this problem 
by building systems to generate personalized reviews.
In other words, we aim to build systems that given a user/item combination, generate the review that the user \emph{would} write, if they reviewed the product.
We show that such systems can generate plausible reviews
that match the style and opinion of a chosen author.
We also show that reviews can be generated 
specifically to reflect a sentiment (star rating) 
or broad category (style of beer).

\begin{figure*}[t]
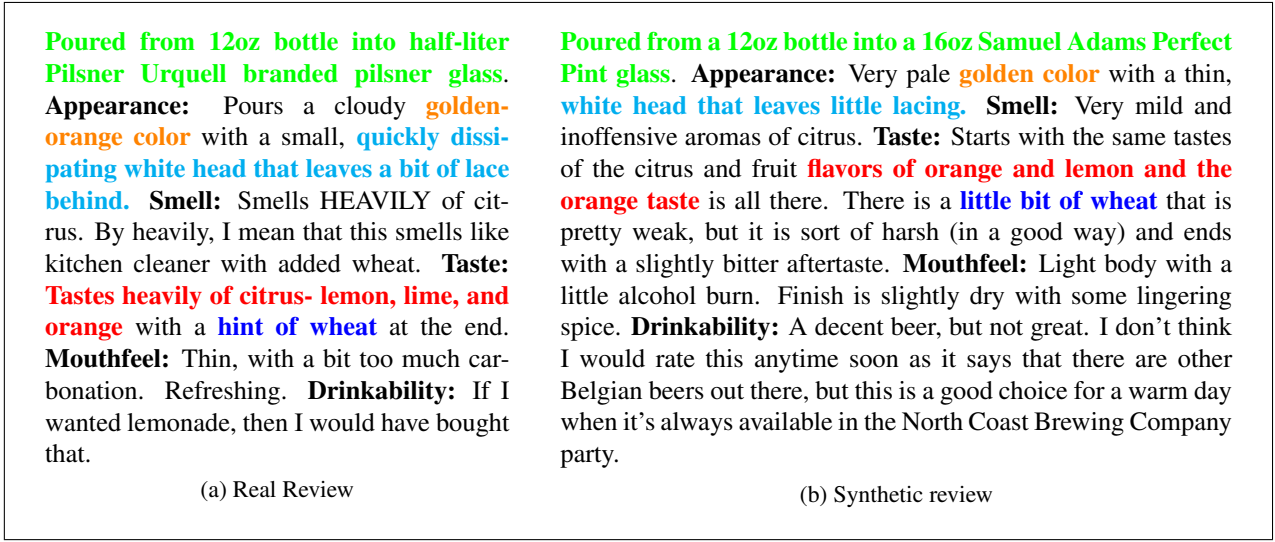

\centering
\fbox{
\begin{subfigure}{.4\linewidth}
\centering
\vphantom{80mm}
\parbox{0.9\linewidth}{
\textbf{\textcolor{green}{Poured from 12oz bottle into half-liter Pilsner Urquell branded pilsner glass}}.  \textbf{Appearance:} Pours a cloudy \textbf{\textcolor{myorange}{golden-orange color}} with a small, \textbf{\textcolor{cyan}{quickly dissipating white head that leaves a bit of lace behind.}} \textbf{Smell:} Smells HEAVILY of citrus. By heavily, I mean that this smells like kitchen cleaner with added wheat.  \textbf{Taste:} \textbf{\textcolor{red}{Tastes heavily of citrus- lemon, lime, and orange}} with a \textbf{\textcolor{blue}{hint of wheat}} at the end.  \textbf{Mouthfeel:} Thin, with a bit too much carbonation. Refreshing.  \textbf{Drinkability:} If I wanted lemonade, then I would have bought that.}
\caption{Real Review}
\vspace{10pt}
\end{subfigure}%
\begin{subfigure}{.56\textwidth}
\centering
\parbox{.93\linewidth}{
\textbf{\textcolor{green}{Poured from a 12oz bottle into a 16oz Samuel Adams Perfect Pint glass}}. \textbf{Appearance:} Very pale \textbf{\textcolor{myorange}{golden color}} with a thin, \textbf{\textcolor{cyan}{white head that leaves little lacing.}} \textbf{Smell:} Very mild and inoffensive aromas of citrus. \textbf{Taste:} Starts with the same tastes of the citrus and fruit \textbf{\textcolor{red}{flavors of orange and lemon and the orange taste}} is all there. There is a \textbf{\textcolor{blue}{little bit of wheat}} that is pretty weak, but it is sort of harsh (in a good way) and ends with a slightly bitter aftertaste. \textbf{Mouthfeel:} Light body with a little alcohol burn. Finish is slightly dry with some lingering spice. \textbf{Drinkability:} A decent beer, but not great. I don't think I would rate this anytime soon as it says that there are other Belgian beers out there, but this is a good choice for a warm day when it's always available in the North Coast Brewing Company party.}
\caption{Synthetic review}
\end{subfigure}
}
\caption{Real and synthetic reviews by user \emph{Mosstrooper} 
describing the beer \emph{Shock Top Belgian White}.
Note that this user and item combination never occurs in the training set.
The GCN captures Mosstrooper's idiosyncracies, 
including his/her tendency to mark each aspect of the beer 
with fully spelled-out headings (\emph{Appearance}, \emph{Smell}, etc., in a particular order).
The GCN also accurately predicts  
that \emph{Mosstrooper} will comment 
on the beer's golden color, citrus flavor, 
and hints of wheat (highlighted in color).
}
\label{fig:author-generate}
\end{figure*}


Our work focuses on reviews scraped from \emph{BeerAdvocate.com} \citep{mcauley2013amateurs}.
Among product review datasets,
BeerAdvocate is notable for its density and for its structure.
The dataset contains thousands of reviewers 
who have written hundreds of reviews each 
and thousands of items that have received hundreds of reviews each.
Notably, reviews exhibit consistent large-scale structure,
discussing  five attributes of each beer (appearance, smell, taste, mouthfeel, and drinkability) in sequence.
BeerAdvocate users exhibit idiosyncratic writing styles, 
adhering to consistent protocols (hyphenation, colons, line-breaks) 
for delineating the various aspects of reviews.
Naturally, the bulk of the variance between users' reviews of a particular beer owes to their own subjective opinions and tastes, something which a successful model must take into account.

Review data also poses subtler challenges.
For example, proper nouns are prominent, 
contributing to an enormous vocabulary, 
and punctuation is essential to the style of reviews.
These are aspects which might be ignored 
or pose computational challenges for word-level models. 

Character-level Recurrent Neural Networks (RNNs)
have a remarkable ability to generate coherent text \citep{sutskever2011generating},
appearing to hallucinate passages 
that plausibly resemble a training corpus.
In contrast to word-level models, 
they do not suffer from computational costs 
that scale with the size of the input or output vocabularies.
Character-level LSTMs powerfully demonstrate the ability of RNNs 
to model sequences on multiple scales simultaneously.
They learn to form words, to form sentences, 
and to generate paragraphs of appropriate length.
To our knowledge, all previous character-level generative models are unsupervised.\footnote{%
Perhaps excepting a manuscript by
\citet{ling2015character}, which describes a machine translation model with a character-level decoder.
}
However, our goal is to generate character-level text 
in a supervised fashion, 
conditioning upon \emph{auxiliary input} 
such as an item's rating or category.%
\footnote{We use \emph{auxiliary input} 
to differentiate the ``context" input 
from the character representation
passed in at each sequence step.
By supervised, we mean the output sequence 
depends upon some auxiliary input.
}
Such conditioning of sequential output 
has been performed successfully with word-level models,
for tasks including machine translation \citep{sutskever2014sequence},
image captioning \citep{vinyals2015show, karpathy2014deep,mao2014deep},
and even video captioning \citep{venugopalan2014translating}.
However, despite the aforementioned virtues 
of character-level models,
no prior work, to our knowledge, 
has successfully trained them in such a supervised fashion.
 
Most supervised approaches to word-level generative text models 
follow the encoder-decoder approach
popularized by \cite{sutskever2014sequence}.
Some auxiliary input, which might be a sentence or an image, 
is encoded by an encoder model as a fixed-length vector.
This vector becomes the initial input to a decoder model,
which then outputs at each sequence step a probability distribution 
predicting the next word.
During training, weights are updated
to give high likelihood 
to the sequences encountered in the training data.
When generating output, 
words are sampled from each predicted distribution 
and passed as input at the subsequent sequence step.
This approach successfully produces coherent and relevant sentences, 
but quality deteriorates as target sequence length exceeds 35 \citep{sutskever2014sequence}.

To model longer passages of text (such as reviews), 
and to do so at the character level, 
we must produce sequences with hundreds or thousands of elements, 
longer than seems practically trainable 
with an encoder-decoder approach.
Attention mechanisms are popular methods
for overcoming the limitations of encoder-decoder architectures \citep{bahdanau2014neural}.
These approaches have two primary aspects. 
First, the input is revisited at each decoding step.
Second, a mechanism is applied to determine which part of the input
to focus on at each step.
This method has been used 
both to perform machine translation \citep{bahdanau2014neural}
and to caption images \citep{xu2015show}.
To overcome the challenges that we face in this work,
we need only the persistence conferred by input replication.
Attention is not required, as our auxiliary information consists of one-hot representations.
Thus we present a stripped down model, 
which we term the Generative Concatenative Network (GCN).
At each sequence step $t$, 
we concatenate the auxiliary input vector $\boldsymbol{x}_{\mathit{aux}}$ 
with the character representation $\boldsymbol{x}_{\mathit{char}}^{(t)}$,
using the resulting vector $\boldsymbol{x'}^{(t)}$ 
to train an otherwise standard generative RNN model.
It might seem redundant to replicate $\boldsymbol{x}_{\mathit{aux}}$ at each sequence step,
but by providing it, we eliminate pressure on the model to memorize it.
Instead, all computation can focus on modeling the text 
and its interaction with the auxiliary input.

\begin{figure}[h]
\centering
\begin{subfigure}{.45\textwidth}
  \centering
  \includegraphics[width=.95\linewidth]{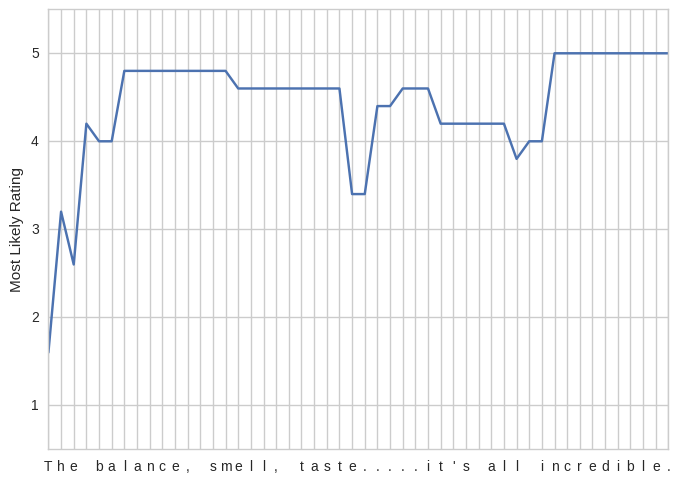}
  \label{fig:incredible}
\end{subfigure}

\begin{subfigure}{.45\textwidth}
  \centering
  \includegraphics[width=.95\linewidth]{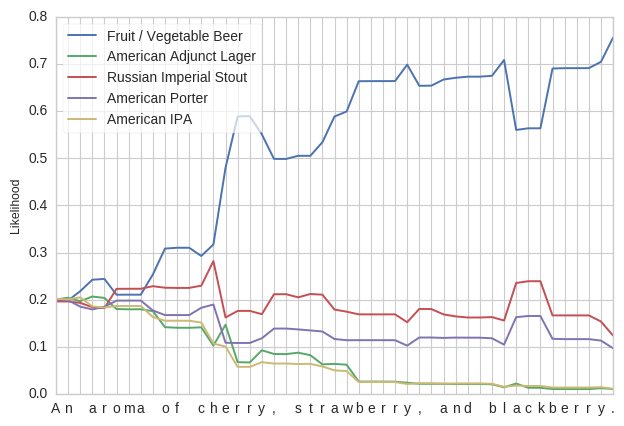}
  \label{fig:fruit}
\end{subfigure}
\vspace{-5pt}
\caption{Our generative model infers ratings and categories given reviews without any \emph{a priori} notion of words.
}
\label{fig:sentiment-teaser}
\vspace{-15pt}
\end{figure} 

We implement a GCN using an LSTM RNN \citep{hochreiter1997long},
demonstrating its efficacy at both review generation 
and traditional supervised learning tasks.
In generative mode, the GCN produces convincing reviews, 
tailored to a star rating and category.
This GCN can also run \emph{in reverse}, 
performing classification with surprising accuracy (\autoref{fig:sentiment-teaser}).
While the chief purpose of our model is to generate text,
we find that classification accuracy of the reverse model 
provides an objective way to assess what the model has learned.
Further, for the task of author identification, 
it rivals the performance of a state-of-the-art 
tf-idf ngram logistic regression classifier.
An empirical evaluation shows that GCNs 
can accurately classify previously unseen reviews as positive or negative 
and determine which of $5$ beer categories is being described,
despite operating at the character level 
and not being optimized directly to minimize classification error.
Our exploratory analysis also reveals 
that the GCN implicitly learns a large vocabulary 
and can effectively model nonlinear dynamics, 
like the effect of negation.
Plotting the inferred rating as each character is encountered 
for many sentences (\autoref{fig:sentiment-teaser})
shows qualitatively that the model infers ratings quickly 
and anticipates words after `reading' particularly informative characters.

%% file: sections/contributions.tex
In this paper we offer the following contributions:
\begin{itemize}
\item A simple character-level RNN architecture 
which effectively generates personalized reviews,
conditioned on the author, item, category, or sentiment. 
\item A demonstration that the generative model can be run \emph{in reverse} as a classifier,
accurately retrieving a review's author or item.
We also show that the same system can classify the sentiment (star rating) of the rating and the category of the item.
\item An empirical evaluation of the generative model showing that the GCN achieves significantly lower perplexity than a standard RNN language model.
\end{itemize}

Beyond the methodology we contribute, 
this work shall help to build novel recommender systems
that offer richer personalization than existing approaches. 
Understanding and summarizing opinion text is a broad topic (see e.g.~\citet{liu04mining}), 
where the goal is often to design a system 
that extracts personalized summaries of reviews (or snippets from reviews) 
that an individual may agree with \citep{lerman09Summ}. 
Our approach attacks this problem more directly, 
generating text that a specific user would be likely to write about a specific product. 
Potentially, our approach coul lead to
more intricately personalized experiences.

%% file: sections/data.tex
We focus on data scraped from BeerAdvocate 
as originally collected and described by \citet{mcauley2013amateurs}.
BeerAdvocate is a large online review community 
boasting 1,586,614 reviews of 66,051 distinct items 
composed by 33,387 users.
The reviews employ a large vocabulary, 
with 241,962 words occurring two times or more 
and 73,394 words occurring 10 times or more.%
\footnote{
This vocabulary exceeds that of \citet{sutskever2014sequence},
who required eight GPUs to train word level models, 
four of which were used to perform softmax across the output layer.
In contrast, we had access to only two GPUs,
and yet our model can train to convergence in several days.
}
Each review is accompanied by a number of numerical ratings, 
corresponding to ``appearance", ``aroma", ``palate", ``taste", and also the user's ``overall" impression.
The reviews are also annotated with the item's category.

In addition to the explicitly structured meta-data (ratings, category, item ID and user ID, etc),
the reviews themselves conform to unenforced protocols.
Typically, each review discusses the \emph{appearance}, \emph{smell}, \emph{taste}, \emph{mouthfeel} and \emph{drinkability} of the beer, 
in precisely this sequence.%
\footnote{\emph{Mouthfeel} refers to the texture of the beer.
\emph{Drinkability} refers to the ease with which the beer can be consumed.
Heavier beers might be tasty but less drinkable while an otherwise despised beer 
(typically American Lagers such as Bud Light) 
could score high points for drinkability, 
even in a review that likens it to urine. 
}
Reviewers frequently demarcate the various sections of the reviews.
One popular style is to abbreviate each section, e.g. in the passage: ``A: Pours a deep amber color with two fingers of foam, leaving little or no lacing on the glass", ``A:" introduces prose describing the beer's appearance.
While some users follow initials with colons (A:, S:, T:, M:, D:), others use hyphens, e.g., ``T- The black cherry is almost completely covered by the alcohol flavor".
Reviewers nearly always adhere to a consistent protocol within a single review.
We have also observed that prolific reviewers
typically adhere to consistent patterns 
across the vast majority of their reviews. 
Thus it is possible, qualitatively, 
to determine whether generated reviews, 
conditioned on a specific author, 
capture that author's style of writing.

For our experiments generating reviews conditioned on rating, user ID, and item ID, 
we select 242k reviews for training and 27k for testing, 
focusing on the most active users and popular items.
We use a standard procedure for selecting dense subgraphs called the $k$-core. 
We simply prune the graph iteratively, 
removing unpopular items and inactive users from the dataset 
until each remaining user has reviewed at least $190$ remaining items and each remaining item has received at least $190$ reviews from remaining users. 
For our experiments focusing on generating reviews conditioned on item category, 
we select a balanced subset consisting of 150k reviews, 
30,000 each from 5 among the top categories,
namely ``American IPA", ``Russian Imperial Stout", ``American Porter", ``Fruit/Vegetable Beer", and ``American Adjunct Lager".
From both datasets, we hold out 10\% of reviews for testing.

%% file: sections/recurrent-nets.tex
Recurrent neural networks extend the capabilities of 
feed-forward networks to handle sequential data.
Inputs $\boldsymbol{x}^{(1)}, ..., \boldsymbol{x}^{(T)}$
are passed to the network one by one.
At each step $t$, the network 
updates its hidden state 
as a function of both the current input
and the previous step's hidden state,
outputting a prediction $\boldsymbol{\hat{y}}^{(t)}$.
In this paper, we use RNNs containing long short term memory (LSTM) cells 
introduced by \cite{hochreiter1997long} with forget gates introduced in \citet{gers2000learning},
owing to their empirical successes and 
demonstrated ability to overcome the exploding/vanishing gradient problems suffered by other RNNs \citep{bengio1994learning}.
In short, each memory cell has an input node $g$, which commonly has a tanh or sigmoid activation function.
The activation from the input node
flows into each cell's internal state $s$,
a structure in which activation is preserved along a self-connected recurrent edge.
Each cell also contains three sigmoidal gating units 
for input ($i$), output ($o$), and to forget ($f$) 
that respectively determine when to let activation (from $g$) into the internal state $s$, 
when to pass activation (from $s$) through to the rest of the network, 
and when to flush the cell's hidden state.
The output of each LSTM layer is another sequence,
allowing us to stack several layers of LSTMs as in \citet{graves2013generating}.
At step $t$, each LSTM layer $\boldsymbol{h}_l^{(t)}$ receives input from the previous layer $\boldsymbol{h}_{l-1}^{(t)}$ at the same sequence step and the same layer at the previous time step $\boldsymbol{h}_{l}^{(t-1)}$. The recursion ends with $\boldsymbol{h}_0^{(t)} = \boldsymbol{x}^{(t)}$ and $\boldsymbol{h}_l^{(0)}=\boldsymbol{0}$.
Formally, for a layer $\boldsymbol{h}_l$ the equations to calculate the forward pass through an LSTM layer are:
$$ \boldsymbol{g}_l^{(t)} = \phi( W_l^{\mbox{gx}} \boldsymbol{h}^{(t)}_{l-1} +   W_l^{\mbox{gh}} \boldsymbol{h}^{(t-1)}_{l}  + \boldsymbol{b}_l^{\mbox{g}})$$
$$ \boldsymbol{i}_l^{(t)}  =    \sigma( W_l^{\mbox{ix}} \boldsymbol{h}^{(t)}_{l-1} + W_l^{\mbox{ih}} \boldsymbol{h}^{(t-1)}_{l} + \boldsymbol{b}_l^{\mbox{i}}) $$
$$ \boldsymbol{f}_l^{(t)}  =    \sigma( W_l^{\mbox{fx}} \boldsymbol{h}^{(t)}_{l-1} + W_l^{\mbox{fh}} \boldsymbol{h}^{(t-1)}_{l} + \boldsymbol{b}_l^{\mbox{f}}) $$
$$ \boldsymbol{o}_l^{(t)} =    \sigma( W_l^{\mbox{ox}} \boldsymbol{h}^{(t)}_{l-1} + W_l^{\mbox{oh}} \boldsymbol{h}^{(t-1)}_{l} + \boldsymbol{b}_l^{\mbox{o}}) $$
$$ \boldsymbol{s}_l^{(t)} = \boldsymbol{g}_l^{(t)} \odot \boldsymbol{i}_l^{(i)} + \boldsymbol{s}_l^{(t-1)} \odot \boldsymbol{f}_l^{(t)}) $$
$$ \boldsymbol{h}^{(t)}_{l}  = \phi(\boldsymbol{s}_l^{(t)}) \odot \boldsymbol{o}_l^{(t)}.$$ 
Here, $\sigma$ denotes an element-wise sigmoid function, $\phi$ an element-wise $tanh$, 
and $\odot$ is an element-wise product.
While a thorough treatment of the LSTM is beyond the scope of this paper, we refer to our review of the literature \citep{lipton2015critical} for a gentler unpacking of the material.

\subsection{Generative RNNs}
\begin{figure}[t]
\centering
  \begin{subfigure}[b]{\linewidth}
  \captionsetup{skip=0pt}
    \centering
      \includegraphics[scale=0.23]{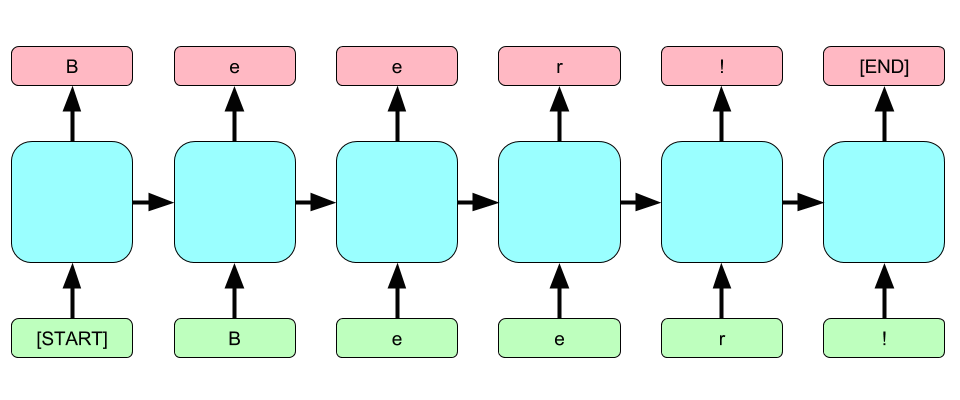}
    \caption{}
    \label{fig:charnet}
  \end{subfigure}
  \vfill
  \begin{subfigure}[b]{\linewidth}
  \captionsetup{skip=5pt}
    \centering
      \includegraphics[scale=0.23]{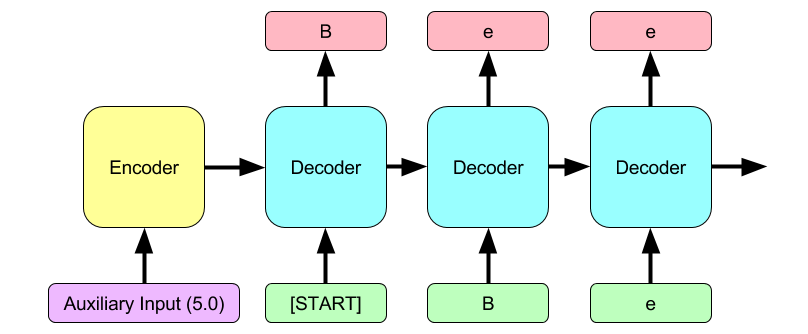}
    \caption{}
    \label{fig:encoder-decoder}
  \end{subfigure}
  \vfill
  \begin{subfigure}[b]{\linewidth}
    \captionsetup{skip=0pt}
    \centering
      \includegraphics[scale=0.23]{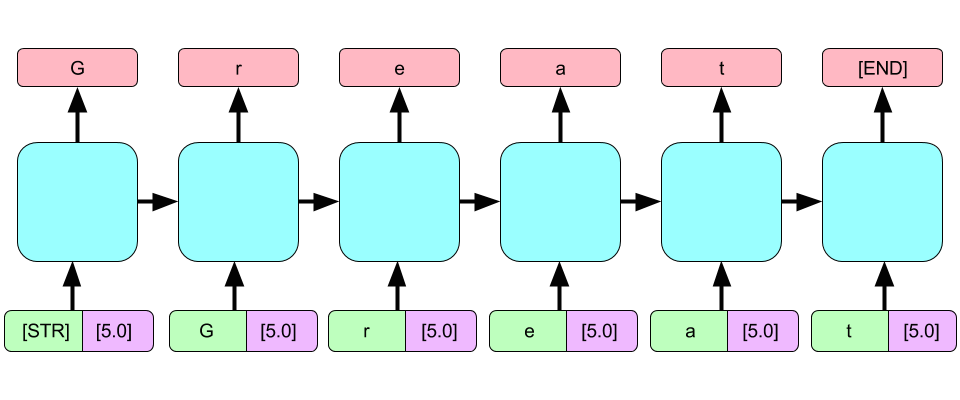}
      \caption{}
    \label{fig:catnet}
  \end{subfigure}%
  \caption{(a) Standard generative RNN; (b) encoder-decoder RNN; (c) concatenated input RNN.}
	\vspace{-15pt}
\end{figure}

Before introducing our contributions,
we review the generative RNN model of \citet{sutskever2011generating, sutskever2014sequence} on which we build.
A generative RNN is trained to predict the next token 
in a sequence, i.e.~$\boldsymbol{\hat{y}}^{t} = \boldsymbol{x}^{(t+1)}$, 
given all inputs to that point ($\boldsymbol{x}^{1},...,\boldsymbol{x}^{t}$). 
Thus input and output strings are equivalent 
but for a one token shift (\autoref{fig:charnet}).
The output layer is fully connected with softmax activation, 
ensuring that outputs specify a distribution.
Cross entropy is the loss function during training. 

Once trained, the model is run in generative mode
by sampling stochastically 
from the distribution output at each sequence step,
given some starting token and state.
Passing the sampled output as the subsequent input,
we  generate another output conditioned on the first prediction,
and can continue in this manner 
to produce arbitrarily long sequences.
Sampling can be done directly according to softmax outputs, 
but it is also common to \emph{sharpen} the distribution 
by setting a \emph{temperature} $\leq 1$, 
analogous to the so-named parameter in a Boltzmann distribution.
Applied to text, generative models trained in this fashion 
produce surprisingly coherent passages 
that appear to reflect the characteristics 
of the training corpus. 
They can also be used to continue passages given some starting tokens. 

\subsection{Generative Concatenative RNNs}
Our goal is to generate text in a supervised fashion, 
conditioned on an auxiliary input $\boldsymbol{x}_{aux}$.
This has been done at the word-level with encoder-decoder models (\autoref{fig:encoder-decoder}),
in which the auxiliary input is encoded and passed as the initial state to a decoder, 
which then must preserve this input signal across many sequence steps \citep{sutskever2014sequence,karpathy2014deep}. 
Such models have successfully produced (short) image captions,
but seem impractical for generating full reviews at the character level because signal from $\boldsymbol{x}_{aux}$ must survive for hundreds of sequence steps.

We take inspiration from an analogy to human text generation.
Consider that given a topic and told to speak at length, 
a human might be apt to meander and ramble. 
But given a subject to stare at, it is far easier to remain focused.
The value of re-iterating high-level material 
is borne out in one study, \citet{Surber2007}, 
which showed that repetitive subject headings in textbooks
resulted in faster learning, less rereading and more accurate answers to high-level questions.

Thus we propose the generative concatenative network,
a simple architecture in which input $\boldsymbol{x}_{aux}$ 
is concatenated with the character representation
$\boldsymbol{x}_{char}^{(t)}$.
Given this new input $\boldsymbol{x'}^{(t)} = [\boldsymbol{x}_{char}^{(t)};\boldsymbol{x}_{aux}]$
we can train the model precisely as with the standard generative RNN (\autoref{fig:catnet}).
At training time, $\boldsymbol{x}_{aux}$ is a feature of the training set.
At prediction time, we fix some $\boldsymbol{x}_{aux}$, 
concatenating it with each character sampled from $\boldsymbol{\hat{y}}^{(t)}$.
One might reasonably note that this replicated input information is redundant.
However, since it is fixed over the course of the review,
we see no reason to require the model 
to transmit this signal across hundreds of time steps.
By replicating $\boldsymbol{x}_{aux}$ at each input, 
we free the model to focus on learning the complex interaction between the auxiliary input and language, rather than memorizing the input.

\subsection{Weight Transplantation}
Models with even modestly sized auxiliary input representations are considerably harder to train than a typical unsupervised character model.
To overcome this problem, we first train a character model to convergence.
Then we transplant these weights into a concatenated input model, initializing the extra weights (between the input layer and the first hidden layer) to zero. 
Zero initialization is not problematic here because symmetry in the hidden layers is already broken. 
Thus we guarantee that the model will achieve a strictly lower loss than a character model, saving (days of) repeated training. 
This scheme bears some resemblance to the pre-training common in the computer vision community \citep{yosinski2014transferable}.
Here, instead of new output weights, we train new input weights.

\subsection{Running the Model in Reverse}
Many common document classification models, like tf-idf logistic regression, maximize the likelihood of the training labels given the text.
Given our generative model, 
we can produce a predictor by reversing the order of inference, 
that is, by maximizing the likelihood of the text, 
given a classification.
The relationship between these two tasks
($P(\boldsymbol{x}_{aux} | \mbox{Review})$ and $P(\mbox{Review} | \boldsymbol{x}_{aux})$)
follows from Bayes' rule.
That is, our model predicts the conditional probability $P(\mbox{Review}| \boldsymbol{x}_{aux})$
of an entire review given some $\boldsymbol{x}_{aux}$ (such as a star rating).
The normalizing term can be disregarded in determining the most probable rating and when the classes are balanced,
as they are in our test cases, 
the prior also vanishes from the decision rule leaving $P(\boldsymbol{x}_{aux}|\mbox{Review}) \propto P(\mbox{Review}|\boldsymbol{x}_{aux})$.
 
 In the specific case of author identification,
 we are solving the problem
 \begin{equation*}
 \begin{split}
 & \argmax_{u \in \textit{users}} P(\mbox{Review}|u) \cdot P(u) \\
= &\argmax_{u} \left( \prod_{t=1}^{T} P(y_t|u, y_1,...,y_{t-1}) \right) \cdot P(u)\\
= & \argmax_{u} \left( \sum_{t=1}^{T} \mbox{log}(P(y_t|u, y_1,...,y_{t-1})) \right) + \mbox{log}\left(P(u)\right)
\end{split}
\end{equation*}
where $y_t$ is the $t^{\mbox{th}}$ character in the review, 
and $T$ is the last character's index.
In practice, we must do these calculations in log space
to avoid losing precision.
The joint probability (over all characters)
of any review is extremely small,
but the relative values of these probabilities
is nevertheless informative.
This method allows us to perform classification 
without any additional training,
and provides an intuitive way to ask,
``what does the generative model know?"
Our experiments show, surprisingly,
that for the task of author identification,
this approach rivals traditional state-of-the-art classifiers. 
It has the benefit that it can make a prediction given a fragment of a document, however short. 
For example, the GCN can be used identify users given the first 100 characters of a review without any modifications to the model.
In contrast, an n-gram logistic regression model would require retraining.
However, one drawback to the model might be 
the inefficiency of separately running each user through the network,
to determine which makes the review most likely.

%% file: sections/experiments.tex
All experiments are executed 
with a custom recurrent neural network library written in Python, 
using Theano \citep{bergstra2010theano} for GPU acceleration. 
Our networks use $2$ hidden layers 
with $1024$ LSTM cells per layer. 
During training, examples are processed in mini-batches 
and we update weights with RMSprop \citep{rmsprop}.
To assemble batches, we concatenate 
all reviews in the training set together, 
delimiting them with ($<$STR$>$)
and ($<$EOS$>$) tokens.
We split this string into mini-batches of size $256$
and again split each mini-batch 
into segments with sequence length $200$. 
Furthermore, LSTM state is preserved across batches during training. 
To combat exploding gradients, 
we clip the elements of each gradient at $\pm$ 5.
We found that we could speed up training 
by first training an unsupervised character-level generative RNN to convergence.
We then transplant weights from the unsupervised net 
to initialize the GCN.
We implement two GCNs in this fashion,
one using the star rating scaled to [-1, 1] as $\boldsymbol{x}_{aux}$, 
and a second using a one-hot encoding of 5 beer categories as $\boldsymbol{x}_{aux}$.

\subsection{Generating Text}
First, we evaluate the reviews generated by the GCN.
Conditioning on authors (\autoref{fig:author-generate}) and items,
we produce reviews that capture the author's style.
This is best appreciated by evaluating side by side
real reviews corresponding to a user-item pair,
and synthesized reviews conditioned on the same user-item pair.
As demonstrated in \autoref{fig:author-generate},
the GCN learns the peculiarities of
user \emph{Mosstrooper}'s style,
demarcating each section in similar fashion.
The GCN also accurately predicts 
several sentiments that \emph{Mosstrooper}
would express regarding \emph{Shock Top Belgian White},
such as its golden color, citrusy flavor, and hints of wheat.
Notably, both our qualitative and quantitative analyses show that user information is far more salient than item information for predicting review text. 

We similarly evaluate the GCN's ability 
to generate reviews 
conditioned on star ratings and categories.
Running the GCN in generative mode 
and conditioning upon a 5 star rating, 
we produce a decidedly positive review:
\begin{displayquote}
Poured from a 12oz bottle into a pint glass.
A: Pours a deep brown color with a thin tan head.
The aroma is of coffee, chocolate, and coffee.
The taste is of roasted malts, coffee, chocolate, and coffee.
The finish is slightly sweet and smooth with a light bitterness and a light bitterness that lingers on the palate.
The finish is slightly bitter and dry.
Mouthfeel is medium bodied with a good amount of carbonation.
The alcohol is well hidden.
Drinkability is good.
I could drink this all day long.
I would love to try this one again and again.
\end{displayquote}

Conditioning on the ``Fruit / Vegetable Beer'' category, the model generates a commensurately botanical review; interestingly the user ``Mikeygrootia'' does not exist in the dataset.
\begin{displayquote}
Thanks to Mikeygrootia for the opportunity to try this one.             
A: Poured a nice deep copper with a one finger head that disappears quickly.  Some lacing.               
S: A very strong smelling beer. Some corn and grain, some apple and lemon peel.      
Taste: A very sweet berry flavor with a little bit of a spice to it. 
I am not sure what to expect from this beer. 
This stuff is a good summer beer. 
I could drink this all day long. Not a bad one for me to recommend this beer.
\end{displayquote}


\subsection{Generative Model Quantitative Results}
To prove that our generative model 
makes use of the auxiliary information 
to produce more contextually likely reviews, we report the test set perplexities
for all models (\autoref{tab:perplexity-results}.
As a baseline, we report the perplexity achieved 
by an unsupervised LSTM language model.
We then report the perplexities of GCNs trained with rating, item ID, and user ID, as auxiliary information.

Because perplexity is a brittle measure, with outliers capable 
of dominating the performance across an entire dataset with unbounded loss, 
we report both the average perplexity over all test set reviews 
and the median perplexity over all test set reviews.

Among all models,
the GCN using user and item information performed best.
User information proved most valuable, and item information conferred little additional power to predict the review. 
In fact, item information ID proved less useful than rating information for explaining the content of reviews.
This struck us as surprising, 
because intuitively, the item gives a decent indication of the rating and fully specifies the beer category. It also suggests which proper nouns are likely to occur in the review.
In contrast, we suspected that user information would be more difficult to use but were surprised that our model could capture user's writing patterns extremely accurately. 
For nearly 1000 users, the model can generate reviews that clearly capture each author's writing style despite forging unique reviews (Figure \ref{fig:author-generate}). 

\begin{table}[t]
\centering
  \begin{tabular}{ l  c c}
\toprule
  \multicolumn{3}{c}{\textbf{Test Set Perplexity}}\\
  \midrule
  & Mean & Median  \\
  Unsupervised Language Model  & 4.23  & 2.22 \\
  Rating  & 2.94   & 2.07  \\
  Item  & 4.48   & 2.17  \\
  User  & 2.26  & 2.03  \\
  User-Item & 2.25   & 1.98  \\
\bottomrule
  \end{tabular}
\caption{Perplexity on test set data for unsupervised RNN language model as well as GCNs with rating, category, user, item, user and item info.}
\label{tab:perplexity-results}
\vspace{-10pt}
\end{table}

\subsection{Predicting Sentiment and Category One Character at a Time}
\begin{figure}[t]
\centering
\begin{subfigure}{.45\textwidth}
  \centering
  \includegraphics[width=.95\linewidth]{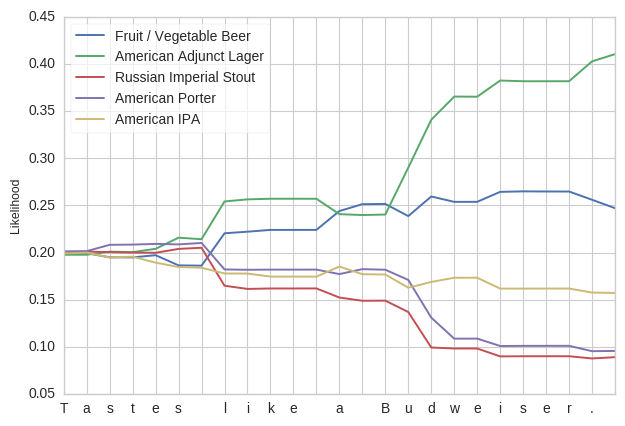}
  \label{fig:tastes-budweiser}
\end{subfigure}
\begin{subfigure}{.45\textwidth}
  \centering
  \includegraphics[width=.95\linewidth]{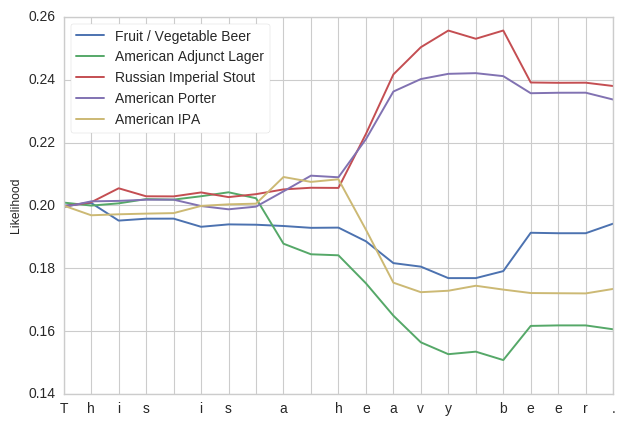}
  \label{fig:aroma-cherry}
\end{subfigure}
\caption{Probability of each category as each character in the review is encountered. The GCN learns Budweiser is a lager 
and that stouts and porters are heavy.}
\label{fig:category-phrasegraph}
\vspace{-10pt}
\end{figure}

\begin{figure}[t]
\begin{subfigure}{.45\textwidth}
  \centering
  \includegraphics[width=.9\linewidth]{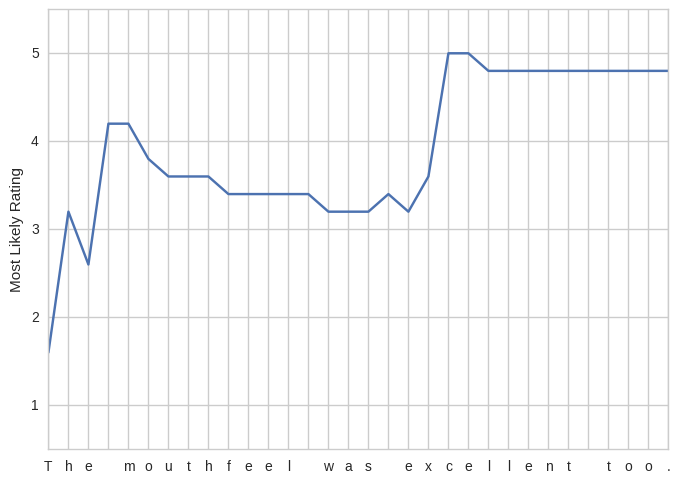}
  \label{fig:excellent}
\end{subfigure}
\begin{subfigure}{.45\textwidth}
  \centering
  \includegraphics[width=.9\linewidth]{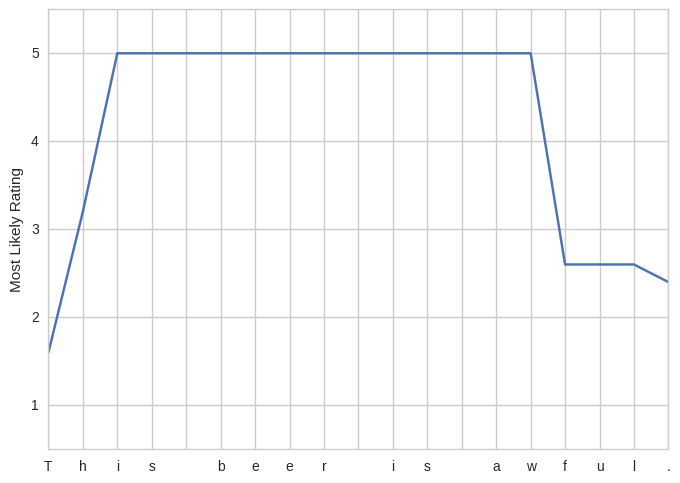}
  \label{fig:awful}
\end{subfigure}
\caption{Most likely star rating
as each letter is encountered.  
The GCN learns to tilt positive by the `c' in `excellent'
and that the `f' in `awful' reveals negative sentiment.}
\label{fig:phrasegraph}
\vspace{-10pt}
\end{figure} 

In addition to running the GCN to generate output, 
we take example sentences from unseen reviews 
and plot the rating which gives the sentence maximum likelihood 
as each character is encountered (\autoref{fig:phrasegraph}).
We can also plot the network's perception of item category,
using each category's prior and the review's likelihood to infer posterior probabilities after reading each character.
These visualizations demonstrate that by the ``d" in ``Budweiser", our model recognizes a ``lager".
Similarly, reading the ``f" in ``awful'', the network seems to comprehend that the beer is ``awful" and not ``awesome'' (\autoref{fig:phrasegraph}).

To verify that the argmax over many settings of the rating is reasonable, we plot the log likelihood after the final character is processed, given by a range of fine-grained values for the rating (1.0, 1.1, etc.). 
These plots show that the log likelihood tends to be smooth,
peaking at an extreme for sentences with unambiguous sentiment, 
e.g., ``Mindblowing experience",
and peaking in the middle
when sentiment is ambiguous
, e.g., ``not the best, not the worst."
We also find that the model understands 
nonlinear dynamics of negation
and can handle simple spelling mistakes.

\subsection{Classification Results}
While our motivation is to produce a character-level generative model, running in reverse-fashion as a classifier proved an effective way to objectively gauge what the model \emph{knows}. 
To investigate this capability more thoroughly, 
we compared it to a word-level tf-idf $n$-gram multinomial logistic regression (LR) model, using the top 10,000 $n$-grams. 
For the task of author identification,
the GCN equals the performance of the ngram tf-idf model (\autoref{tab:confusion-mat}). 
However, classifying items proved more difficult.
On the task of category prediction,
the GCN achieves a classification accuracy 
of $89.9\%$ while LR achieves $93.4\%$ (\autoref{tab:confusion-category}). 
Both models make the majority of their mistakes confusing Russian Imperial Stouts for American Porters, 
which is not surprising because stouts 
are a sub-type of porter. 
If we collapse these two into one category, 
the RNN achieves $94.7\%$ accuracy 
while LR achieves $96.5\%$.
While the reverse model does not in this case eclipse a state of the art classifier, 
it was trained at the character level and was not optimized to minimize classification error or with attention to generalization error.
In this light, the results appear to warrant a deeper exploration of this capability. 
We also ran the model in reverse to classify results as positive ($\geq 4.0$ stars) or negative ($\leq 2.0$ stars), achieving AUC of $.88$ on a balanced test set with $1000$ examples (\autoref{tab:confusion-rat-ratnet}).

\begin{table}[ht]
\setlength{\tabcolsep}{5pt}
\centering
  \begin{tabular}{ c c c c }
  \toprule
  \multicolumn{4}{c}{\textbf{Predicting User from Review}}\\
   & Accuracy & AUC & Recall@10\%  \\
\midrule
   GCN-Character & .9190  & .9979  & .9700 \\
   TF-IDF n-gram & .9133 & .9979  & .9756 \\
\midrule
  \multicolumn{4}{c}{\textbf{Predicting Item from Review}}\\
   & Accuracy & AUC & Recall@10\%  \\
\midrule
     GCN-Character & .1280 & .9620  & .4370 \\
   TF-IDF n-gram & .2427 & .9672  & .5974 \\
\bottomrule
  \end{tabular}
  \setlength{\tabcolsep}{6pt}
\caption{Information retrieval performance for predicting the author of a review and item described.}
\label{tab:confusion-mat}
\vspace{-10pt}
\end{table}


\begin{table}[ht]
\begin{subtable}{\linewidth}
\centering
  \begin{tabular}{c  c c c c c c}
\toprule
  & & \multicolumn{5}{c}{\textbf{Predicted}}\\
  & & F/V & Lager & Stout & Porter & IPA \\
\midrule
  \multirow{5}{*}{\textbf{True}} 
  & F/V & 910 & 28  & 7  & 14  & 41 \\
  & Lager & 50  & 927 & 3   & 3  & 17 \\
  & Stout & 16   & 1   & 801 & 180 & 2  \\
  & Porter & 22   & 3   & 111  & 856 & 8\\
  & IPA & 19  & 12   & 4   & 12  & 953\\
\bottomrule
  \end{tabular}
  \caption{GCN}
 \end{subtable}
 \begin{subtable}{\linewidth}
 \centering
  \begin{tabular}{c c c c c c c}
\toprule
  & & \multicolumn{5}{c}{\textbf{Predicted}}\\
  & & F/V & Lager & Stout & Porter & IPA \\
\midrule
  \multirow{5}{*}{\textbf{True}} & F/V & 923 & 36 & 9 & 10 & 22\\
  & Lager & 16 & 976 & 0 & 1 & 7 \\
  & Stout & 9 & 4 & 920 & 65 & 2 \\
  & Porter & 11 & 6 & 90 & 887 & 6\\
  & IPA & 18 & 13 & 1 & 2 & 966 \\
\bottomrule
  \end{tabular}
\caption{ngram tf-idf.}
 \end{subtable}
\caption{Beer category classification confusion matrices.}
\vspace{-10pt}
\label{tab:confusion-category}
\end{table}

\begin{table}[t]
	\centering
  \begin{tabular}{c c c c}
\toprule
  \multirow{2}{*}{\textbf{Method}} & \multirow{2}{*}{\textbf{True Label}} & \multicolumn{2}{c}{\textbf{Predicted Label}}\\
& & Negative & Positive \\
\midrule
  \multirow{2}{*}{\textbf{GCN}} & Negative &  294 & 206 \\
  & Positive & 7 & 493\\
  \midrule
    \multirow{2}{*}{\textbf{ngram tf-idf}} & Negative &  459 & 41 \\
  & Positive & 42 & 458\\
\bottomrule
  \end{tabular}
\caption{Sentiment (rating) classification confusion matrices for GCN (top) and ngram tf-idf (bottom).}
\label{tab:confusion-rat-ratnet}
\vspace{-15pt}
\end{table}

\subsection{Learning Nonlinear Dynamics of Negation}
\begin{figure}[h]
\centering
\begin{subfigure}{.5\linewidth}
  \centering
  \includegraphics[width=1\linewidth]{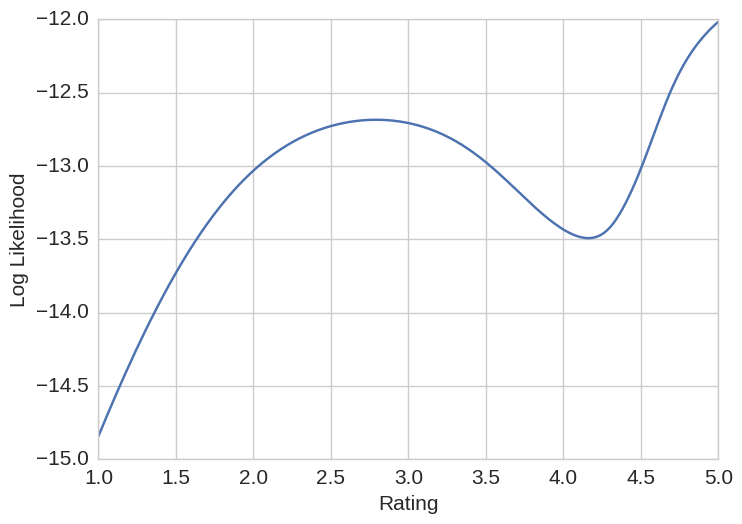}
  \caption{``This beer is great."}
  \label{fig:great}
\end{subfigure}%
\begin{subfigure}{.5\linewidth}
  \centering
  \includegraphics[width=1\linewidth]{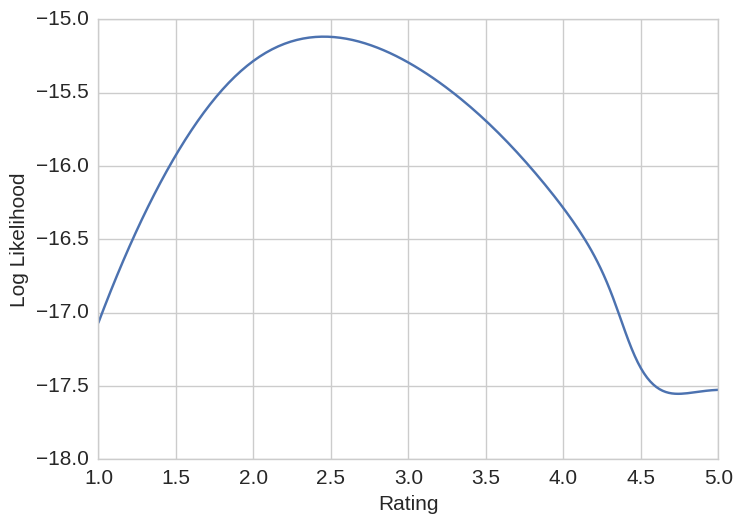}
  \caption{``This beer is not great"}
  \label{fig:not-great}
\end{subfigure}
\begin{subfigure}{.5\linewidth}
  \centering
  \includegraphics[width=1\linewidth]{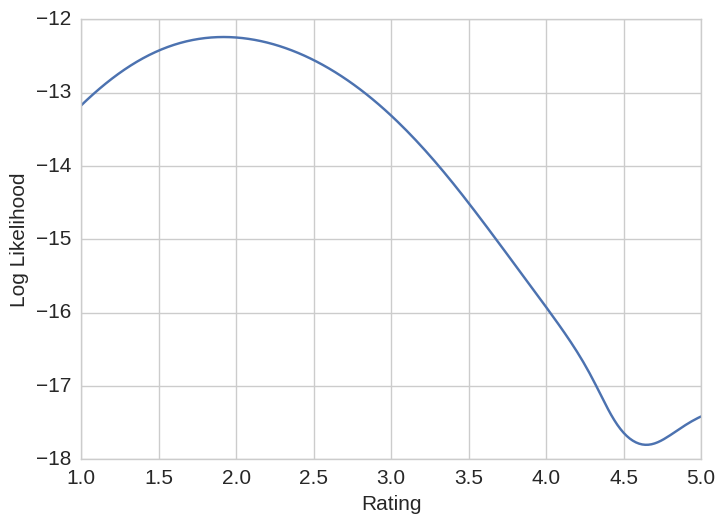}
  \caption{``This beer is bad."}
  \label{fig:bad}
\end{subfigure}%
\begin{subfigure}{.5\linewidth}
  \centering
  \includegraphics[width=1\linewidth]{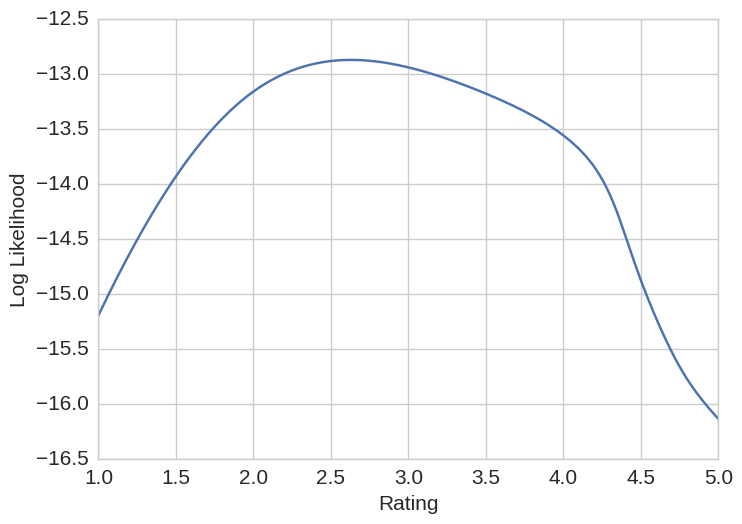}
  \caption{``This beer is not bad."}
  \label{fig:not-bad}
\end{subfigure}
\caption{We plot the likelihood given 
to a review by each rating. The network learns nonlinear dynamics of negation. ``Not" reduces the rating when applied to ``great" but increases the rating when applied to ``bad". }
\label{fig:negation-phrasegraph}
\vspace{-10pt}
\end{figure}

By qualitatively evaluating the beliefs of the network about the sentiment (rating) 
corresponding to various phrases,
we can easily show many clear cases where the GCN is able to model the nonlinear dynamics in text.
To demonstrate this capacity,
we plot the likelihoods of the review 
conditioned on each setting of the rating for the phrases ``this beer is great",
``this beer is not great",
``this beer is bad",
and ``this beer is not bad'' \autoref{fig:negation-phrasegraph}.

%% file: sections/related.tex
The prospect of capturing meaning in character-level text 
has long captivated neural network researchers.
In the seminal work, ``Finding Structure in Time",
\citet{elman1990finding} speculated,
``one can ask whether the notion `word'
(or something which maps on to this concept) 
could emerge as a consequence of learning 
the sequential structure of letter sequences 
that form words and sentences 
(but in which word boundaries are not marked)."
In this work, an `Elman RNN' was trained with $5$ input nodes, 
$5$ output nodes,
and a single hidden layer of $20$ nodes,
each of which had a corresponding context unit to predict the next character in a sequence. 
At each step, 
the network received a binary encoding (not one-hot) 
of a character and tried to predict 
the next character's binary encoding.
Elman plots the error of the net character by character,
showing that it is typically high at the onset of words,
but decreasing as it becomes clear what each word is.
While these nets do not possess the size or capabilities of large modern LSTM networks trained on GPUs, this work lays the foundation for much of our research.
Subsequently, in 2011,
\citet{sutskever2011generating}
introduced the model of text generation 
on which we build. 
In that paper, the authors 
generate text resembling Wikipedia articles and New York Times articles. 
They sanity check the model by showing 
that it can perform a \emph{debagging} task
in which it unscrambles bag-of-words representations of sentences 
by determining which unscrambling has the highest likelihood. 
Also relevant to our work is \citet{zhang2015text},
which trains a strictly discriminative model of text at the character level using convolutional neural networks \citep{lecun1989backpropagation,lecun1998gradient}.
Demonstrating success on both English and Chinese language datasets,
their models achieve high accuracy on a number of classification tasks.
\citet{dai2015semi} train a character level LSTM to perform document classification,
using LSTM RNNs pretrained as either language models or 
sequence-to-sequence auto-encoders.

Related work generating sequences in a supervised fashion 
generally follow the pattern of \citet{sutskever2014sequence},
which uses a word-level encoder-decoder RNN to map sequences onto sequences.
Their system for machine translation demonstrated that a recurrent neural network can compete with state of the art machine translation systems 
absent any hard-coded notion of language (beyond that of words).
Several papers followed up on this idea, extending it to image captioning 
by swapping the encoder RNN for a convolutional neural network \citep{mao2014deep, vinyals2015show, karpathy2014deep}.
Most similar to our generative model, \cite{bahdanau2014neural}
introduced the idea of an attention mechanism for the task of machine translation.
In this model, the entire input is revisited at each decoding step.
The attention mechanism determines which part of the input
to focus on at each step.
This idea was subsequently revisited in the context of machine translation 
by \citet{xu2015show}, who apply attention to the task of image captioning. 
While attention has been increasingly well-studied, 
to our knowledge no papers have simplified the model,
studying generation with replicated inputs independent of the complex machinery of attention.

\subsection{Key Differences and Contributions}
RNNs have been used previously to generate text at the character level.
And they have been used to generate text in a supervised fashion at the word-level.
However, to our knowledge, this is the first work to demonstrate that an RNN 
can generate relevant text at the character level.
Further, while \cite{sutskever2011generating} 
demonstrates the use of a character level RNN as a scoring mechanism 
for the toy problem of unscrambling strings,
to our knowledge, this is the first paper to use generative model likelihood scores to infer labels, simultaneously learning to generate text 
and to perform classification with high accuracy.
Our work is not the first to demonstrate a character-level classifier, as \citet{zhang2015text} offered such an approach. 
However, while their model is strictly discriminative, 
our model's main purpose is to generate text, 
a capability missing from their approach.
Further, while we present a preliminary exploration of ways that our generative model can be used as a classifier,  
we do not train it directly to minimize classification error or generalization error, rather using the classifier interpretation to validate that the generative model is in fact modeling the auxiliary information meaningfully.

%% file: sections/discussion.tex
In this work, we demonstrate 
a \emph{character-level} 
recurrent neural network to generate relevant text conditioned on auxiliary input.
It is the first attempt to generate personalized product reviews,
conditioned on specific users and items with deep learning.
It is also the first work
to use deep learning for generating coherent product reviews conditioned upon rating and category data. 
The conditionally generated reviews achieve significantly lower mean perplexity 
than those achieved with standard RNN language models.

Our quantitative and qualitative analysis 
show that the GCN can accurately identify authors. 
Additionally, the model does a good job of predicting items ratings and categories, nearly matching the performance of purely discriminative models.
The classification accuracy of the generative model provides a straightforward way to determine what the model \emph{knows}. 
Unlike mean perplexity, 
the classification results 
are less susceptible to outliers.
The GCN learns nonlinear dynamics of negation,
and appears to respond intelligently to a large vocabulary
despite lacking any \emph{a priori} notion of words.

While the capability of the generative model to perform classification is intriguing, 
more work can be done to tune this approach.
One problem with the current approach is that inference becomes slow as the number of classes becomes large.
In the case of author identification, 
we must run each review through the network roughly 1k times
to obtain a likelihood of the review
separately conditioned upon each of the roughly 1k authors. 
At scale, this might not be unacceptable. 
How best to perform this form of inference with a deep net (recovering most probable inputs given outputs) remains an interesting and open research question.

%% file: generative-concatenative.bbl
\begin{thebibliography}{26}
\providecommand{\natexlab}[1]{#1}
\providecommand{\url}[1]{\texttt{#1}}
\expandafter\ifx\csname urlstyle\endcsname\relax
  \providecommand{\doi}[1]{doi: #1}\else
  \providecommand{\doi}{doi: \begingroup \urlstyle{rm}\Url}\fi

\bibitem[Bahdanau et~al.(2014)Bahdanau, Cho, and Bengio]{bahdanau2014neural}
Dzmitry Bahdanau, Kyunghyun Cho, and Yoshua Bengio.
\newblock Neural machine translation by jointly learning to align and
  translate.
\newblock \emph{arXiv preprint arXiv:1409.0473}, 2014.

\bibitem[Bengio et~al.(1994)Bengio, Simard, and Frasconi]{bengio1994learning}
Yoshua Bengio, Patrice Simard, and Paolo Frasconi.
\newblock Learning long-term dependencies with gradient descent is difficult.
\newblock \emph{Neural Networks, IEEE Transactions on}, 5\penalty0
  (2):\penalty0 157--166, 1994.

\bibitem[Bergstra et~al.()Bergstra, Breuleux, Bastien, Lamblin, Pascanu,
  Desjardins, Turian, Warde-Farley, and Bengio]{bergstra2010theano}
James Bergstra, Olivier Breuleux, Fr{\'e}d{\'e}ric Bastien, Pascal Lamblin,
  Razvan Pascanu, Guillaume Desjardins, Joseph Turian, David Warde-Farley, and
  Yoshua Bengio.
\newblock Theano: A cpu and gpu math compiler in python.

\bibitem[Dai and Le(2015)]{dai2015semi}
Andrew~M Dai and Quoc~V Le.
\newblock Semi-supervised sequence learning.
\newblock In \emph{Advances in Neural Information Processing Systems}, pages
  3061--3069, 2015.

\bibitem[Elman(1990)]{elman1990finding}
Jeffrey~L. Elman.
\newblock Finding structure in time.
\newblock \emph{Cognitive science}, 14\penalty0 (2):\penalty0 179--211, 1990.

\bibitem[Gers et~al.(2000)Gers, Schmidhuber, and Cummins]{gers2000learning}
Felix~A. Gers, J{\"u}rgen Schmidhuber, and Fred Cummins.
\newblock Learning to forget: {C}ontinual prediction with {LSTM}.
\newblock \emph{Neural {c}omputation}, 12\penalty0 (10):\penalty0 2451--2471,
  2000.

\bibitem[Graves(2013)]{graves2013generating}
Alex Graves.
\newblock Generating sequences with recurrent neural networks.
\newblock \emph{arXiv preprint arXiv:1308.0850}, 2013.

\bibitem[Hochreiter and Schmidhuber(1997)]{hochreiter1997long}
Sepp Hochreiter and J{\"u}rgen Schmidhuber.
\newblock Long short-term memory.
\newblock \emph{Neural {C}omputation}, 9\penalty0 (8):\penalty0 1735--1780,
  1997.

\bibitem[Hu and Liu(2004)]{liu04mining}
Minqing Hu and Bing Liu.
\newblock Mining and summarizing customer reviews.
\newblock In \emph{KDD}, 2004.

\bibitem[Karpathy and Fei-Fei(2014)]{karpathy2014deep}
Andrej Karpathy and Li~Fei-Fei.
\newblock Deep visual-semantic alignments for generating image descriptions.
\newblock \emph{arXiv preprint arXiv:1412.2306}, 2014.

\bibitem[LeCun et~al.(1989)LeCun, Boser, Denker, Henderson, Howard, Hubbard,
  and Jackel]{lecun1989backpropagation}
Yann LeCun, Bernhard Boser, John~S Denker, Donnie Henderson, Richard~E Howard,
  Wayne Hubbard, and Lawrence~D Jackel.
\newblock Backpropagation applied to handwritten zip code recognition.
\newblock \emph{Neural {c}omputation}, 1\penalty0 (4):\penalty0 541--551, 1989.

\bibitem[LeCun et~al.(1998)LeCun, Bottou, Bengio, and
  Haffner]{lecun1998gradient}
Yann LeCun, L{\'e}on Bottou, Yoshua Bengio, and Patrick Haffner.
\newblock Gradient-based learning applied to document recognition.
\newblock \emph{Proceedings of the IEEE}, 86\penalty0 (11):\penalty0
  2278--2324, 1998.

\bibitem[Lerman et~al.(2009)Lerman, Blair-Goldensohn, and
  McDonald]{lerman09Summ}
Kevin Lerman, Sasha Blair-Goldensohn, and Ryan McDonald.
\newblock Sentiment summarization: Evaluating and learning user preferences.
\newblock In \emph{ACL}, 2009.

\bibitem[Ling et~al.(2015)Ling, Trancoso, Dyer, and Black]{ling2015character}
Wang Ling, Isabel Trancoso, Chris Dyer, and Alan~W Black.
\newblock Character-based neural machine translation.
\newblock \emph{arXiv preprint arXiv:1511.04586}, 2015.

\bibitem[Lipton et~al.(2015)Lipton, Berkowitz, and Elkan]{lipton2015critical}
Zachary~C. Lipton, John Berkowitz, and Charles Elkan.
\newblock A critical review of recurrent neural networks for sequence learning.
\newblock \emph{arXiv preprint arXiv:1506.00019}, 2015.

\bibitem[Mao et~al.(2014)Mao, Xu, Yang, Wang, and Yuille]{mao2014deep}
Junhua Mao, Wei Xu, Yi~Yang, Jiang Wang, and Alan Yuille.
\newblock Deep captioning with multimodal recurrent neural networks (m-{RNN}).
\newblock \emph{arXiv preprint arXiv:1412.6632}, 2014.

\bibitem[McAuley and Leskovec(2013)]{mcauley2013amateurs}
Julian~John McAuley and Jure Leskovec.
\newblock From amateurs to connoisseurs: modeling the evolution of user
  expertise through online reviews.
\newblock In \emph{Proceedings of the 22nd international conference on World
  Wide Web}, pages 897--908. International World Wide Web Conferences Steering
  Committee, 2013.

\bibitem[Surber and Schroeder(2007)]{Surber2007}
John~R. Surber and Mark Schroeder.
\newblock {Effect of Prior Domain Knowledge and Headings on Processing of
  Informative Text}.
\newblock \emph{Contemporary Educational Psychology}, 32\penalty0 (3):\penalty0
  485--498, jul 2007.
\newblock ISSN 0361476X.
\newblock \doi{10.1016/j.cedpsych.2006.08.002}.
\newblock URL
  \url{http://www.sciencedirect.com/science/article/pii/S0361476X06000348}.

\bibitem[Sutskever et~al.(2011)Sutskever, Martens, and
  Hinton]{sutskever2011generating}
Ilya Sutskever, James Martens, and Geoffrey~E. Hinton.
\newblock Generating text with recurrent neural networks.
\newblock In \emph{Proceedings of the 28th International Conference on Machine
  Learning (ICML-11)}, pages 1017--1024, 2011.

\bibitem[Sutskever et~al.(2014)Sutskever, Vinyals, and
  Le]{sutskever2014sequence}
Ilya Sutskever, Oriol Vinyals, and Quoc~V. Le.
\newblock Sequence to sequence learning with neural networks.
\newblock In \emph{Advances in Neural Information Processing Systems}, pages
  3104--3112, 2014.

\bibitem[Tieleman and Hinton(2012)]{rmsprop}
Tijmen Tieleman and Geoffrey~E. Hinton.
\newblock Lecture 6.5- {RMSprop}: Divide the gradient by a running average of
  its recent magnitude.
\newblock \url{https://www.youtube.com/watch?v=LGA-gRkLEsI}, 2012.

\bibitem[Venugopalan et~al.(2014)Venugopalan, Xu, Donahue, Rohrbach, Mooney,
  and Saenko]{venugopalan2014translating}
Subhashini Venugopalan, Huijuan Xu, Jeff Donahue, Marcus Rohrbach, Raymond
  Mooney, and Kate Saenko.
\newblock Translating videos to natural language using deep recurrent neural
  networks.
\newblock \emph{arXiv preprint arXiv:1412.4729}, 2014.

\bibitem[Vinyals et~al.(2015)Vinyals, Toshev, Bengio, and
  Erhan]{vinyals2015show}
Oriol Vinyals, Alexander Toshev, Samy Bengio, and Dumitru Erhan.
\newblock Show and tell: A neural image caption generator.
\newblock In \emph{Proceedings of the IEEE Conference on Computer Vision and
  Pattern Recognition}, pages 3156--3164, 2015.

\bibitem[Xu et~al.(2015)Xu, Ba, Kiros, Courville, Salakhutdinov, Zemel, and
  Bengio]{xu2015show}
Kelvin Xu, Jimmy Ba, Ryan Kiros, Aaron Courville, Ruslan Salakhutdinov, Richard
  Zemel, and Yoshua Bengio.
\newblock Show, attend and tell: Neural image caption generation with visual
  attention.
\newblock \emph{arXiv preprint arXiv:1502.03044}, 2015.

\bibitem[Yosinski et~al.(2014)Yosinski, Clune, Bengio, and
  Lipson]{yosinski2014transferable}
Jason Yosinski, Jeff Clune, Yoshua Bengio, and Hod Lipson.
\newblock How transferable are features in deep neural networks?
\newblock In \emph{Advances in Neural Information Processing Systems}, pages
  3320--3328, 2014.

\bibitem[Zhang and LeCun(2015)]{zhang2015text}
Xiang Zhang and Yann LeCun.
\newblock Text understanding from scratch.
\newblock \emph{arXiv preprint arXiv:1502.01710}, 2015.

\end{thebibliography}
